\documentclass[runningheads]{llncs}
\usepackage[T1]{fontenc}
\usepackage{amsmath}
\usepackage{graphicx}
\usepackage{multirow}
\usepackage{marvosym}
\usepackage{makecell}
\usepackage{enumitem}
\usepackage[colorlinks,bookmarksopen,bookmarksnumbered,citecolor=blue, linkcolor=blue, urlcolor=blue]{hyperref}

\begin{document}
\title{Vision-Based Localization in Dense Urban Environments: A Case Study of an Urban Village in China}
\titlerunning{Vision-Based Localization in Dense Urban Environments}

\author{Menglin Wu\thanks{The first author is a full-time student.} \and
Rui Cao$^{(\textrm{\Letter})}$}
\authorrunning{M. Wu and R. Cao}

\institute{Thrust of Urban Governance and Design, Society Hub, The Hong Kong
University of Science and Technology (Guangzhou), Guangzhou, China\\
\email{ruicao@hkust-gz.edu.cn}}
\maketitle
\begin{abstract}
Urban villages, the widespread informal settlements which have emerged as a result of rapid urbanization, are now major residential hubs for migrant workers in large cities in China. The dense arrangement of buildings in these areas often leads to unreliable GPS signals, while incomplete mapping data further impairs accurate route planning and navigation. These issues not only hinder everyday mobility but also pose significant challenges for emergency response, as confusing road layouts and GPS inaccuracies can complicate evacuation efforts. To address these challenges, we propose a practical vision-based geo-localization solution tailored for dense urban environments. Our approach features a low-cost data collection pipeline utilizing a dual-camera system, comprising a panoramic camera and a smartphone camera, to capture synchronized 360-degree panoramas and query images. Using Shipai Village, a well-known densely populated urban village in Guangzhou, as a case study, we develop a specialized image geo-localization dataset. We then assess and compare the performance of existing models across various scene types to identify their strengths and weaknesses. The findings demonstrate both the potential and limitations of visual-based localization in dense urban-village environments.  Our framework aims to enhance pedestrian navigation, last-mile delivery, and emergency management in areas with poor GPS coverage, ultimately supporting the vulnerable populations living within these informal settlements.

\keywords{Vision-based localization \and Image Geo-localization \and Street View Imagery \and Urban Villages \and Sustainable Development Goals (SDGs) \and AI for Social Good}
\end{abstract}

\section{Introduction}
\begin{figure}[t]
    \centering
    \includegraphics[width=0.9\linewidth]{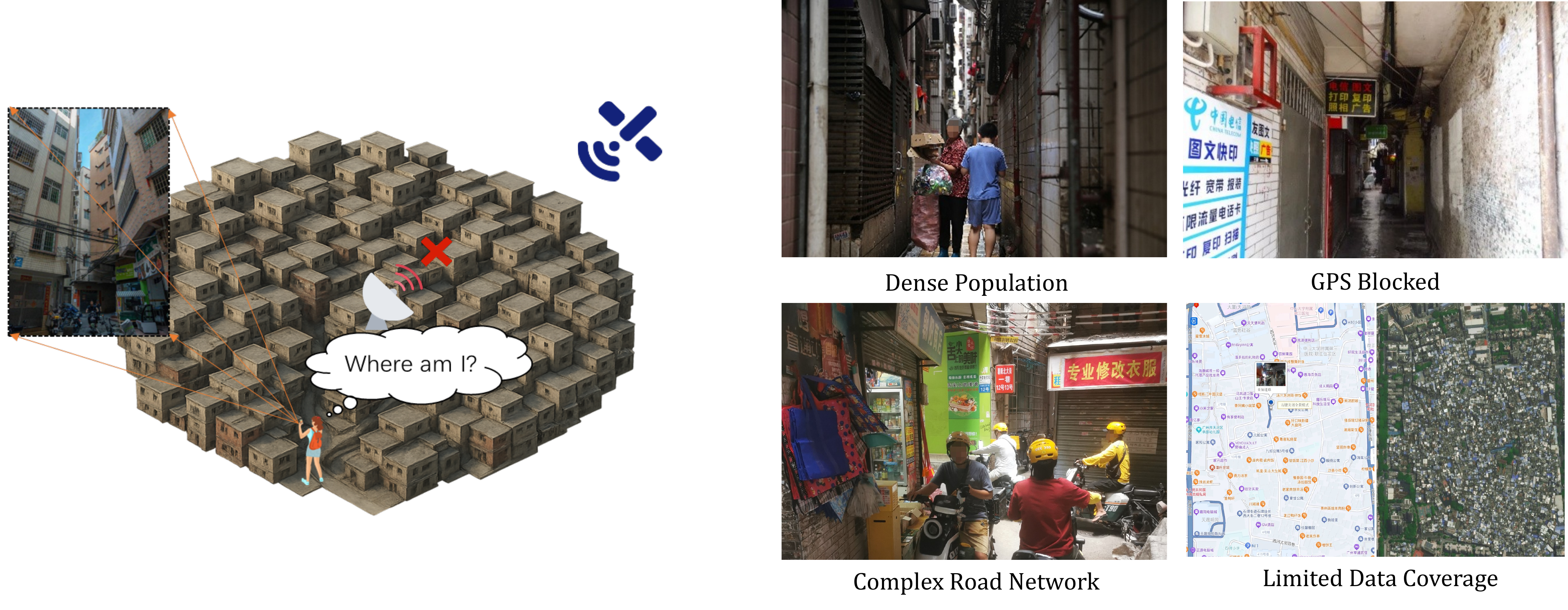}
    \caption{Unique challenges posed by urban villages for reliable geo-localization.}
    \label{fig:challenges}
\end{figure}

Accurate geo-localization underpins a wide range of urban activities, from everyday navigation and logistics to emergency response and public safety. In most modern cities, satellite-based positioning systems combined with detailed digital maps provide reliable support for these tasks. However, such systems are not applicable across all urban spaces. In particular, informal high-density settlements, commonly known as urban villages, expose fundamental limitations of existing localization technologies due to incomplete mapping coverage, the scarcity of reliable fine-grained geospatial data~\cite{cao2025mapping}, and frequent degradation of GPS signals, as illustrated in Fig. \ref{fig:challenges}.

Urban villages emerge when former rural settlements are gradually enclosed by expanding cities \cite{zhang2025mapping}. Over time, land is intensively developed to accommodate growing populations, leading to compact building arrangements, narrow alleys, and irregular circulation patterns. Unlike planned neighborhoods, these areas lack standardized road hierarchies and consistent building orientations. Sky visibility is often severely restricted, and pedestrian movement occurs through fine-grained passages that are difficult to observe from aerial platforms. As a result, conventional GPS-based localization becomes unreliable, and digital maps struggle to reflect the true navigable space. 

The United Nations' Sustainable Development Goals (SDGs) guide the global urban development toward greater inclusiveness, safety, resilience, and sustainability. Among them, SDG 11~\cite{un_sdg11_web} explicitly aims to “make cities and human settlements inclusive, safe, resilient, and sustainable,” with particular emphasis on equitable access to services, safe and efficient mobility, and improved emergency preparedness in densely populated areas. However, in urban villages, these objectives are challenged by extremely high population densities and constrained spatial layouts, where limited public space and insufficient social distancing further increase vulnerability to cascading risks during everyday activities and emergency situations.

The dense and complex spatial structure of urban villages directly affects human mobility. Residents primarily navigate using local visual cues rather than formal street names or mapped routes, making movement decisions at the scale of doorways, staircases, and small intersections. For individuals unfamiliar with the area, the lack of reliable localization support often leads to disorientation and inefficient movement, where even small localization errors can substantially disrupt pedestrian mobility, logistics efficiency, and access to essential services.

The consequences of inaccurate localization extend beyond daily inconvenience and become critical in emergency situations. For instance, when an incident occurs inside an urban village, responders may reach the surrounding area quickly but face difficulty locating the exact entrance or pathway to the affected site. Visually similar facades, hidden access points, and ambiguous building boundaries can delay response even when approximate coordinates are known. In time-sensitive scenarios, such as fires or medical emergencies, these delays can have severe implications. This underscores that localization in urban villages is not merely a positioning problem, but a key factor influencing safety and resilience in densely populated communities.

Image geo-localization~\cite{berton2022deep} provides a promising direction for addressing these challenges. By estimating location directly from visual content, it aligns more closely with how humans perceive and navigate complex spaces. However, existing research in image geo-localization primarily focuses on structured urban streets~\cite{torii2013visual,torii201524,ali2022gsv,berton2022rethinking}, landmarks~\cite{weyand2016planet,weyand2020google}, or aerial–ground~\cite{workman2015wide,liu2019lending} image matching under relatively open conditions. Dense informal settlements, where visual repetition, occlusion, low-illumination, and extreme viewpoint constraints are prevalent, remain largely underrepresented in current benchmarks. This lack of representative datasets hinders systematic evaluation of model robustness and failure modes under fine-scale, GPS-denied conditions, while the reliance on expensive sensing platforms or carefully aligned mapping resources further limits scalability in informal settlements.

Motivated by these gaps, this work investigates image geo-localization in urban villages from a human-centered perspective. In summary, our contributions are threefold:

\begin{itemize}[leftmargin=12pt]
    \item We introduce a novel and low-cost data acquisition and processing pipeline specifically designed to facilitate efficient and reliable data collection in highly complex urban environments, such as dense urban villages. This framework addresses practical constraints while ensuring scalability and robustness.
    \item To the best of our knowledge, we present the first image geo-localization dataset tailored for dense urban villages—a setting characterized by extreme visual challenges, including repetitive architectural patterns, low illumination, motion-induced blur and severe occlusion. Unlike existing benchmarks that focus on structured streets or cross-view settings, our dataset captures highly constrained pedestrian-scale scenarios. The dataset fills a critical gap in existing resources and provides a benchmark for advancing research in geo-localization under real-world urban conditions.
    \item We conduct a systematic evaluation of widely adopted retrieval-based image geo-localization approaches, offering an in-depth analysis of their performance in dense urban village scenarios. Our findings reveal both the strengths and inherent limitations of current methodologies, thereby identifying key directions for future research.
\end{itemize}

The rest of the paper is organized as follows. Section \ref{sec:related-work} reviews the progress of methods and datasets of image geo-localization. Section \ref{sec:method} elaborates the study area and methods. Section \ref{sec:experiments} introduces the experimental settings. Section \ref{sec:results} presents the results and analysis of the experiments. Finally, Section \ref{sec:conclusion} concludes the paper.

\section{Related Work} \label{sec:related-work}
\subsection{Image Geo-localization Methods}
Image geo-localization aims to estimate the geographic location of a query image by matching its visual content against a geo-referenced database. Early approaches relied on hand-crafted local descriptors such as SIFT~\cite{lowe2004distinctive} and SURF~\cite{bay2006surf}, typically combined with Bag-of-Words or VLAD-based encodings for large-scale image retrieval. While effective under limited appearance variation, these methods struggled with extreme viewpoint changes, illumination differences, and dynamic occlusions.

Recently, significant progress in this field has been driven by advances in deep learning and the increasing availability of large-scale geo-tagged imagery datasets~\cite{weyand2016planet,weyand2020google,vo2017revisiting,warburg2020mapillary,berton2022rethinking,ali2022gsv}. The advent of convolutional neural networks (CNNs) marked a major shift toward learning-based representations. NetVLAD~\cite{arandjelovic2016netvlad} introduced an end-to-end trainable aggregation layer that significantly improved place recognition performance by learning compact global descriptors. Subsequent works such as DELF~\cite{noh2017large} explored attention-based local feature selection, while DOLG~\cite{yang2021dolg} combined global and local representations to enhance discriminability across challenging conditions. CosPlace~\cite{berton2022rethinking} targets large-scale scalability by casting training as a classification problem, enabling compact descriptors with strong cross-domain robustness.

More recently, transformer-based architectures and hybrid designs have further improved robustness to viewpoint and appearance variations. TransVPR~\cite{wang2022transvpr} leverages multi-level Transformer attention to aggregate task-relevant regions into global descriptors and further supports re-ranking using patch-level cues. R$^2$Former~\cite{zhu2023r2former} unifies retrieval and re-ranking within a Transformer framework by modeling cross-image correlations, attention signals, and spatial coordinates for more reliable place recognition. In parallel, visual foundation model features have enabled training-free or weakly supervised VPR pipelines. AnyLoc~\cite{keetha2023anyloc} combines self-supervised ViT backbones~\cite{caron2021emerging,oquab2023dinov2} with unsupervised aggregation. It extracts patch-level features from intermediate Transformer layers and aggregates them using VLAD-style pooling, yielding strong generalization across diverse environments without task-specific fine-tuning.

Beyond descriptor learning, recent research has focused on improving feature aggregation strategies and viewpoint robustness. Ge et al.~\cite{ge2020self} proposed mining image-to-region similarities under weak GPS supervision, enabling fine-grained region-level matching across difficult positive pairs. MixVPR~\cite{ali2023mixvpr} introduces a holistic aggregation module that uses MLP-based feature mixing to model global relationships across backbone feature maps and aggregates them into a single global descriptor, improving robustness while remaining lightweight. SALAD~\cite{izquierdo2024optimal} revisits global descriptor construction by formulating feature-to-cluster assignment as an optimal transport problem, enabling selective suppression of non-informative local features via a dustbin mechanism. This design produces compact and informative descriptors while maintaining a single-stage retrieval pipeline. SuperVLAD~\cite{lu2024supervlad} and CricaVPR~\cite{lu2024cricavpr} further refine how local features are aggregated and correlated across images, achieving strong performance with compact descriptors suitable for large-scale deployment. Collectively, these methods push the frontier of image geo-localization by enhancing discriminability, compactness, and robustness under diverse environmental conditions.

\subsection{Geo-localization Datasets and Evaluation Benchmarks}
The development of image geo-localization methods has been closely linked to the availability of benchmark datasets, which differ substantially in viewpoint, sensing modality, and acquisition strategy. Existing datasets can be broadly categorized into cross-view benchmarks and ground-view datasets collected using vehicle-mounted sensors at street level.

Cross-view datasets focus on matching images captured from different viewpoints, typically between ground-level and aerial imagery. Benchmarks such as CVUSA~\cite{workman2015wide} and CVACT~\cite{liu2019lending} have played a pivotal role in advancing cross-view geo-localization by enabling large-scale evaluation of ground-to-aerial matching. These datasets emphasize viewpoint invariance and domain adaptation, but are generally collected in relatively open or structured environments, where road networks and building layouts are clearly observable from above.

Street-level datasets based on vehicle-mounted sensors constitute the most widely used benchmarks for retrieval-based geo-localization. Early large-scale datasets such as Pittsburgh250k~\cite{torii2013visual} and Google Landmarks~\cite{weyand2016planet,weyand2020google} provide extensive street-view imagery with broad geographic coverage, supporting systematic evaluation of large-scale retrieval methods. Subsequent benchmarks, including Tokyo 24/7~\cite{torii201524} and Nordland~\cite{sunderhauf2013we}, introduce additional challenges related to illumination changes, seasonal variation, and long-term appearance drift. More recently, MSLS~\cite{warburg2020mapillary} have expanded data diversity through crowd-sourced, vehicle-based imagery. Despite their scale and diversity, these datasets primarily capture scenes from wide roads and structured urban streets, where camera viewpoints are elevated and motion trajectories are relatively regular.

Pedestrian-scale datasets for micro-level urban environments remain comparatively underexplored. Existing benchmarks are rarely designed to capture narrow passages, dense building layouts, and frequent occlusions that characterize informal settlements (as shown in Fig. \ref{fig:challenges}). Moreover, many datasets rely on vehicle-mounted cameras, professional surveying platforms, or carefully aligned map resources, which entail high equipment costs and limit scalability in densely populated or hard-to-access areas. As a result, localization performance at the pedestrian level, particularly in GPS-degraded environments, has not been sufficiently evaluated.

Recent studies have begun to acknowledge the importance of geo-localization in fine-grained and GPS-denied urban settings~\cite{ye2024coarse}. However, systematic datasets tailored to pedestrian navigation in dense urban informal settlements, such as urban villages, are still scarce. This gap restricts the assessment of model robustness under conditions that are critical for human mobility, emergency response, and safety in high-density communities, motivating the need for dedicated datasets that reflect these real-world challenges.

\section{Materials and Methods} \label{sec:method}
\subsection{Study Area}
\begin{figure}[h]
    \centering
    \includegraphics[width=0.9\linewidth]{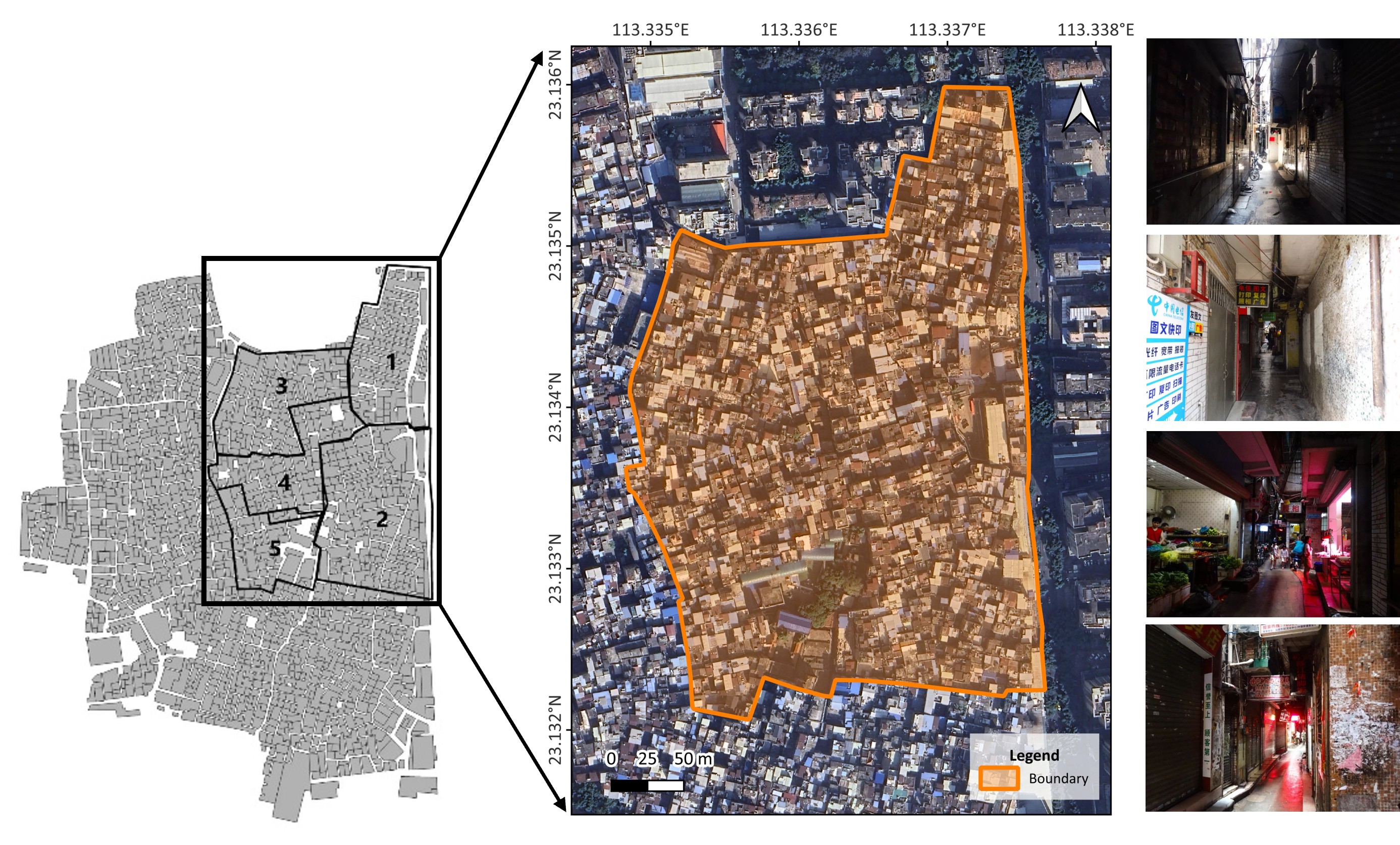}
    \caption{The study area in Shipai Village, Guangzhou, China. Left: Map of Shipai Village. Middle: Satellite view of the study area. Right: Street views within the study area.}
    \label{fig:study_area}
\end{figure}

The study area is located in the Chao Yang community in the northeastern part of Shipai village, one of the most representative surviving urban villages in Tianhe District, Guangzhou, China. The selected region covers approximately $0.1$ km$^2$, as shown in Fig.~\ref{fig:study_area}. 
This informal settlement was formed when a once rural settlement was surrounded by rapid urban development. Over time, local residents built dense rental housing to serve migrant workers and low-income residents. As a result, Shipai today is a very compact neighborhood with extremely narrow alleys, irregular building layouts, limited public space and a mix of old and new structures, as shown in the satellite and street view images in Fig.~\ref{fig:study_area}. 

The complex built form of Shipai causes serious navigation problems. The road network inside the village is like a maze, with many dead ends, tiny passages and overlapping paths. Dense buildings on both sides of the lanes block the sky, so GPS signals are unstable and often jump to wrong positions. On digital maps, many internal paths are missing or simplified, so a planned route on the map may be impossible to follow in reality. Newcomers, delivery couriers and visitors easily get lost. For residents, this makes daily life less convenient, but the issue goes beyond inconvenience. In emergencies, such as a fire or sudden illness, it can be hard to find the exact doorway in time. The neighborhood struggles with safe, accessible public space, and reliable basic infrastructure.

In this work, we leverage the characteristics of Shipai village to construct a specialized image geo-localization dataset, consisting of synchronized panoramic imagery and smartphone query images collected along its narrow alleys. This real-world setting enables us to test the practicality of our proposed low-cost data collection pipeline and to perform a comprehensive assessment of model robustness under dense, irregular, and GPS-degraded urban conditions.

\subsection{Data Collection and Processing}
\begin{figure}[h]
    \centering
    \includegraphics[width=0.8\linewidth]{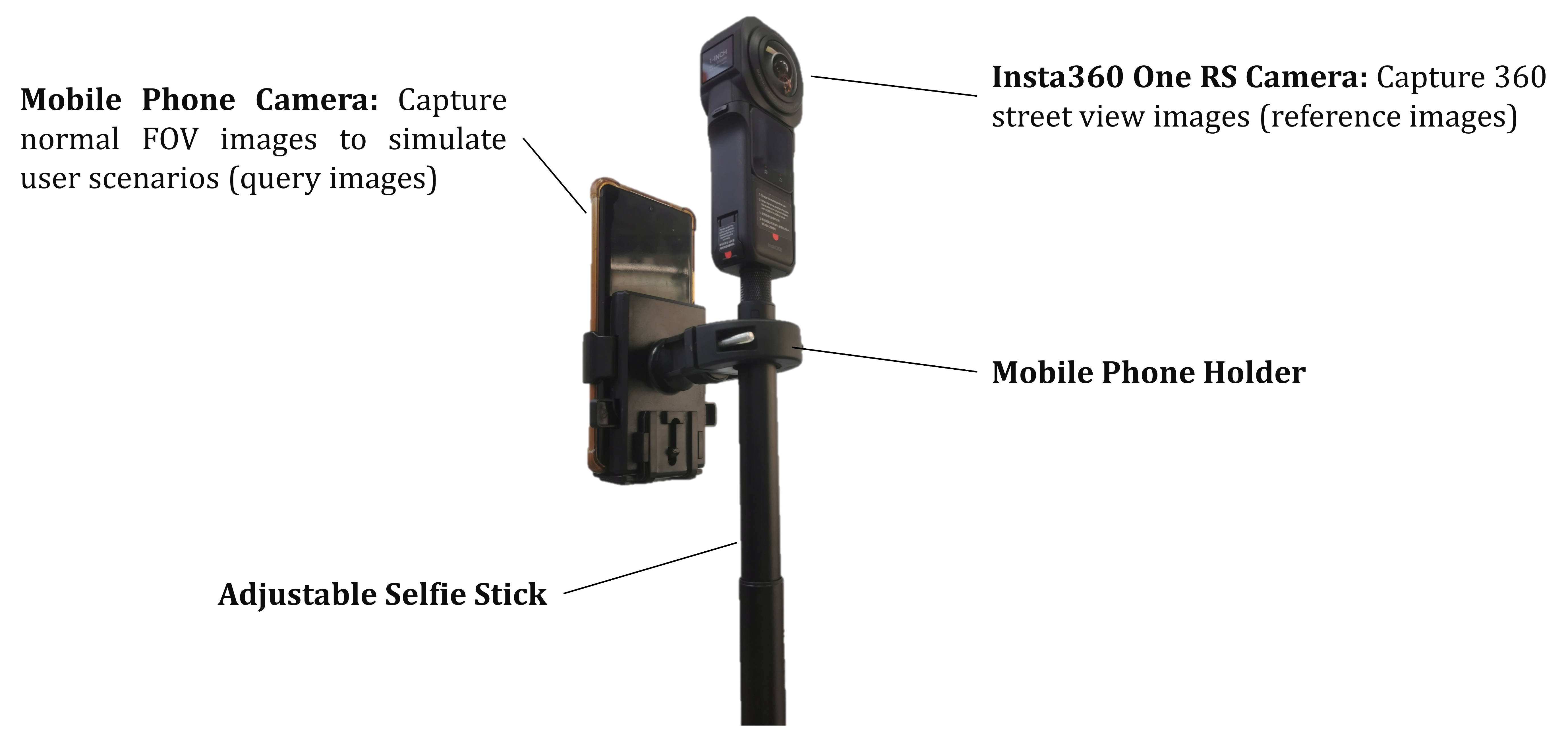}
    \caption{Dual-camera data collection setup. 
    The Insta360 One~RS camera records $360^\circ$ panoramic street-view images to build 
    the reference database, while the smartphone camera captures normal FOV images to 
    simulate user queries. Both cameras are mounted on an adjustable selfie stick and 
    synchronized at 30\,FPS.}
    \label{fig:data_collect}
\end{figure}

To construct a real-world dataset that is both practical and scalable for dense urban informal settlements, we designed a dual-camera data acquisition system consisting of an Insta360 One RS panoramic camera and a mobile phone camera, as shown in Fig.~\ref{fig:data_collect}. Traditional surveying instruments, such as total stations or 3D laser scanners, are typically expensive, heavy, and labor-intensive to deploy, making them impractical for large-scale data collection in narrow and densely populated urban villages. In contrast, our system leverages lightweight, consumer-grade devices that are easy to carry and operate, significantly reducing both equipment cost and manpower requirements.

\begin{figure}[h]
    \centering
    \includegraphics[width=\linewidth]{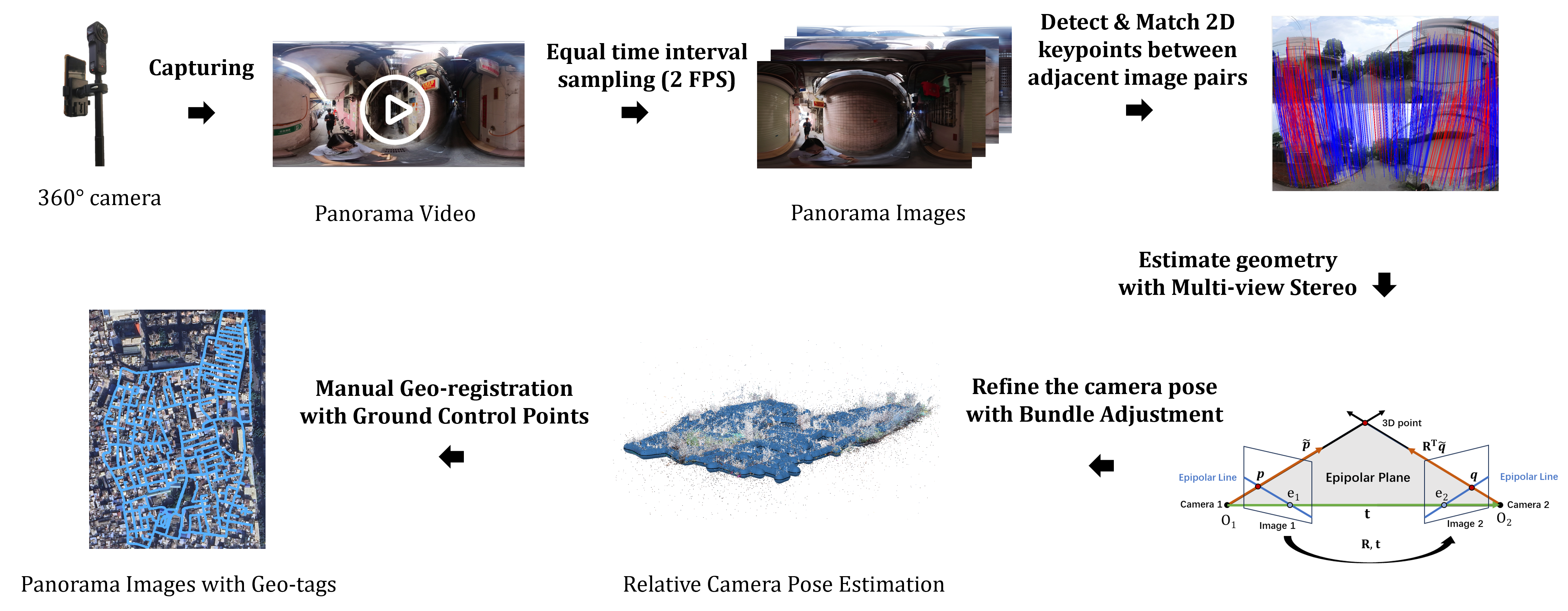}
    \caption{Overall data processing workflow for generating geo-registered panoramic reference images.}
    \label{fig:data_processing}
\end{figure}

Specifically, the Insta360 device was used to collect $360^\circ$ panoramic video frames, which serve as the reference image set for building the retrieval database, while the mobile phone camera was used to capture forward-facing video frames as query images. Both devices were recorded simultaneously at $30 \, \text{FPS}$ and synchronized using timestamp alignment.

The data pre-processing workflow is presented in Fig.~\ref{fig:data_processing}. First, the raw videos collected were temporally sampled at a rate of 2 frames per second to obtain a sequence of images. For each pair of adjacent panoramic frames, 2D keypoints were detected and matched, allowing the relative motion of the camera to be estimated through multi-view geometry and subsequently refined by bundle adjustment. The resulting relative camera pose estimations were then geo-referenced manually with a set of ground control points (GCPs), enabling geographic coordinates to be assigned to all panoramic frames. These geo-registered panoramic images serve as the foundation for the following geo-localization experiments.

To reduce the geometric distortion that typically appears near the edges of $360^\circ$ panoramic images and to improve the quality of visual feature extraction, each panoramic frame was further converted into a set of perspective images. Specifically, every panorama was sliced into $8$ perspective views using a $45^\circ$ heading interval, a $60^\circ$ field of view (FOV), and a $5^\circ$ pitch offset. This process ensures full directional coverage when building the reference image database and enables multi-view matching during retrieval. 

\begin{figure}[h]
    \centering
    \includegraphics[width=\linewidth]{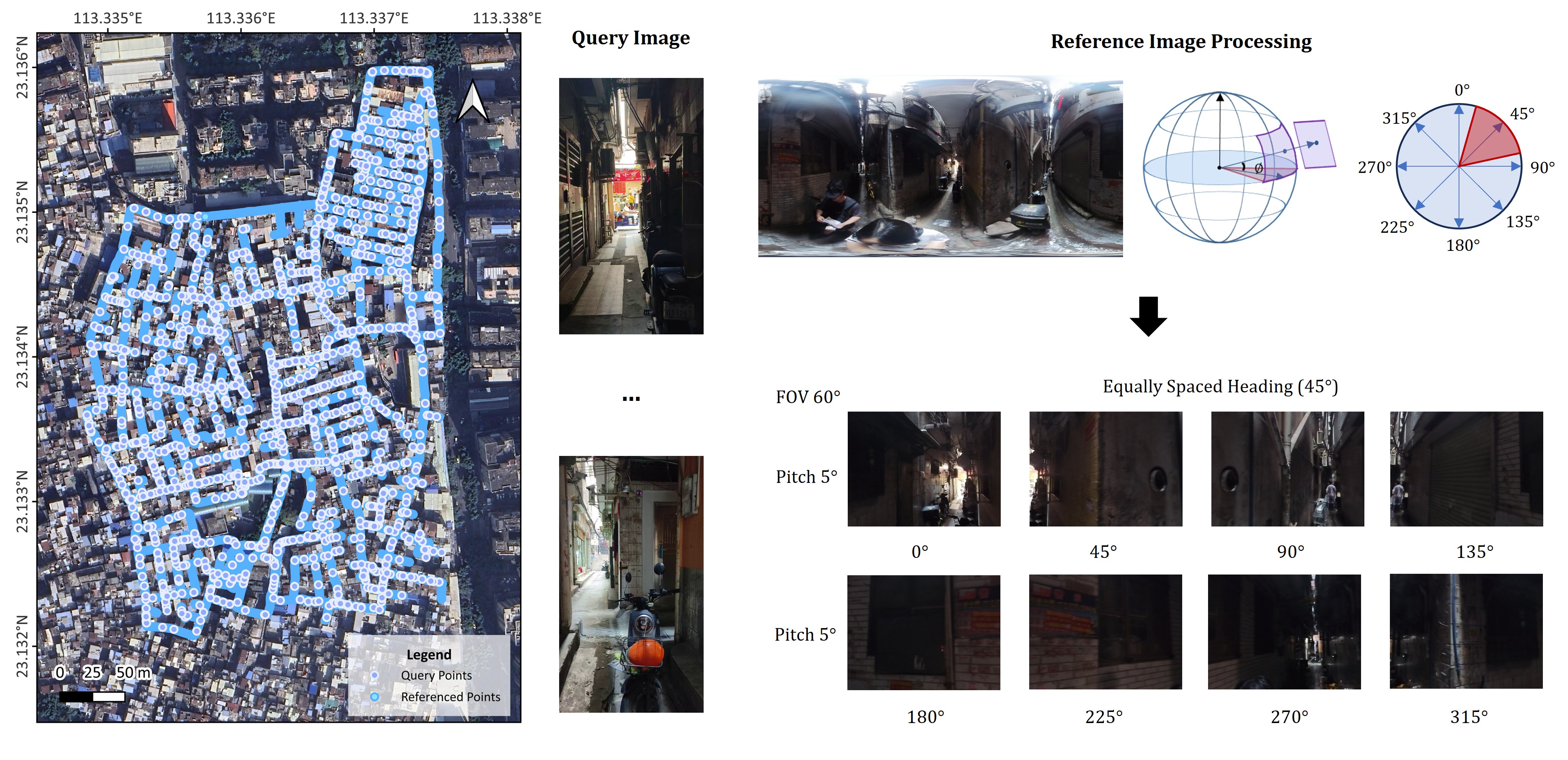}
    \caption{The spatial distribution of the query and reference images in the ShipaiVillage dataset.}
    \label{fig:data_distribution}
\end{figure}

The resulting dataset, which will be referred to as \textit{ShipaiVillage dataset} in the following, contains 228,304 reference images and 1417 query images, forming a benchmark for evaluating image geo-localization in dense urban environments. The spatial distribution of the reference and query images is shown in Fig.~\ref{fig:data_distribution}.
It should be noted that this dataset allows controlled evaluation of retrieval-based localization performance under real-world challenges including motion blur, lighting variation, occlusion, and high visual similarity across scene locations.

\subsection{Problem Definition}
The problem can be formally expressed as follows. Given a query image $I_q$ captured by users in the target environment, the objective is to estimate its location by retrieving the most visually similar geo-referenced image from a pre-constructed database:
\begin{equation}
D = \{(I_1, L_1), (I_2, L_2), \dots, (I_n, L_n)\},
\label{eq:database}
\end{equation}
where each $I_i$ is a reference image and $L_i$ is its associated ground-truth location.

Let $f(\cdot)$ denote a feature extraction model that encodes an image into a feature embedding, and let $\text{dist}(\cdot,\cdot)$ be a similarity metric (e.g., L2 distance, inner product, or cosine similarity). The predicted location $\hat{L_q}$ is obtained by retrieval:
\begin{equation}
\hat{L_q} = L_r \quad \text{where} \quad r = \arg\min_{i} \; \text{dist}\big(f(I_q), f(I_i)\big).
\label{eq:retrieval}
\end{equation}

\subsection{Overall Pipeline}
\begin{figure}[h]
    \centering
    \includegraphics[width=\linewidth]{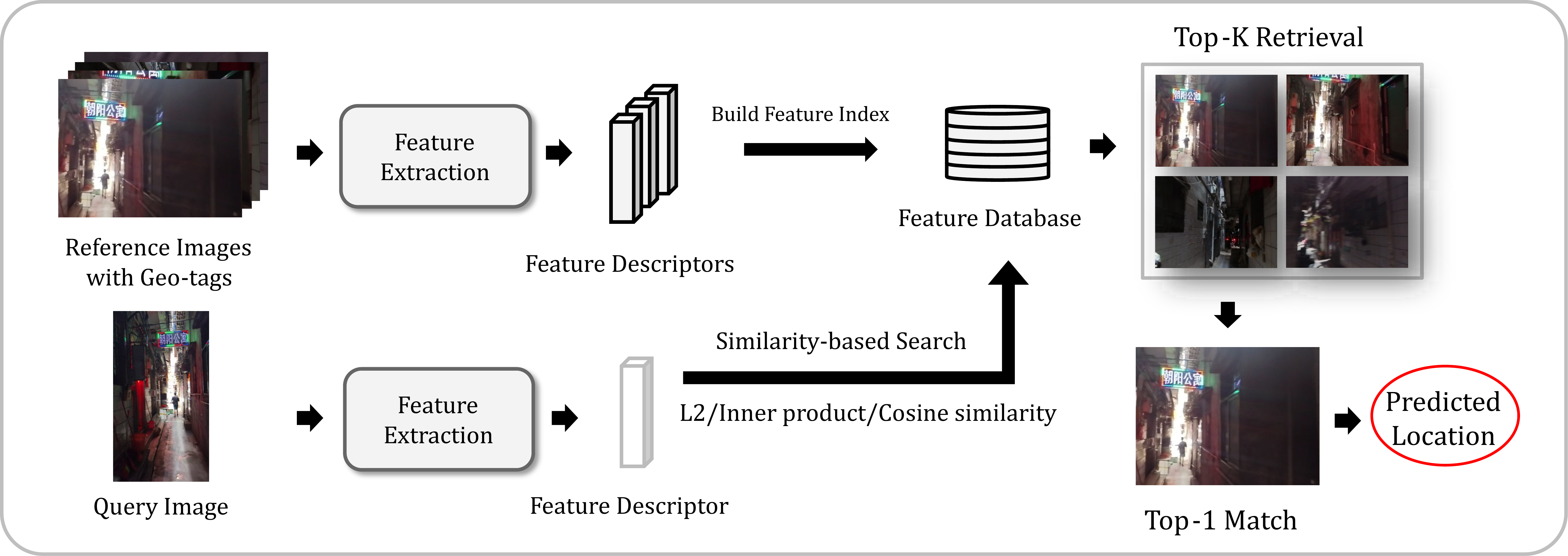}
    \caption{Overview of the retrieval-based image geo-localization pipeline.}
    \label{fig:pipeline}
\end{figure}

The overall workflow follows a retrieval-based image geo-localization pipeline, as shown in Fig.~\ref{fig:pipeline}. The system consists of two parallel stages: reference feature database construction and online query localization.

In the offline stage, a set of geo-tagged reference images is first processed by a feature extraction model to extract global feature descriptors. These descriptors are then indexed to form a feature database, enabling efficient large-scale similarity search.

During inference, a query image captured by a user (e.g., from a mobile phone) is passed through the same feature extractor to obtain its embedding. The query descriptor is matched against the reference feature database using a similarity metric such as L2 distance, inner product, or cosine similarity. The system returns the Top-K most similar reference images, from which the Top-1 match is selected and its associated geo-coordinate is assigned as the predicted location.

\section{Experiments} \label{sec:experiments}
\subsection{Evaluation Metrics}
To provide a comprehensive evaluation of geo-localization performance, we measure positional accuracy using two widely adopted distance-based metrics: Average Error Distance (AED) and Median Error Distance (MED). These metrics measure the geodesic discrepancy between the predicted location and the ground-truth position of each query image.

Given a set of $N$ query samples, let $\mathbf{p}_i = (\phi_i, \lambda_i)$ denote the predicted latitude and longitude for the $i$-th query, and $\mathbf{g}_i = (\phi^{}_i, \lambda^{}_i)$ its ground-truth coordinates. The localization error for each sample is computed as the geodesic distance $d(\mathbf{p}_i, \mathbf{g}_i)$ between the predicted and ground-truth coordinate pairs, measured as the shortest path along the Earth’s surface on the WGS-84 ellipsoidal model that accounts for the Earth’s curvature.

\begin{itemize}
    \item \textbf{Average Error Distance (AED)}: captures the mean localization error across all queries, reflecting overall performance:
    \begin{equation}
        \mathrm{AED} = \frac{1}{N} \sum_{i=1}^{N} d(\mathbf{p}_i, \mathbf{g}_i).
    \end{equation}

    \item \textbf{Median Error Distance (MED)}: provides the median localization error, offering robustness to outliers:
    \begin{equation}
        \mathrm{MED} = \mathrm{median}\left( d(\mathbf{p}_i, \mathbf{g}_i) \right).
    \end{equation}
\end{itemize}

\subsection{Implementation Details}
Here we describe the implementation details as follows. We conducted all experiments on a workstation equipped with an NVIDIA RTX 4090 GPU. To ensure a fair comparison, both query and reference images were resized to a fixed resolution of $322{\times}322$ and normalized without applying any augmentations that alter the appearance. We evaluated multiple representative image geo-localization models, including SFRS~\cite{ge2020self}, CosPlace~\cite{berton2022rethinking}, Conv-AP~\cite{ali2022gsv}, MixVPR~\cite{ali2023mixvpr}, 
EigenPlaces~\cite{berton2023eigenplaces},
SALAD~\cite{izquierdo2024optimal}, CricaVPR~\cite{lu2024cricavpr} and SuperVLAD~\cite{lu2024supervlad} encoders, using their official pretrained weights. Each model produces a global feature descriptor, computed in mixed precision (FP16) to accelerate inference and reduce memory usage. All reference descriptors were L2-normalized and indexed using FAISS. During inference, each query image was passed through the same encoder to obtain its descriptor, which was then matched against the feature index to retrieve the Top-K nearest neighbors, from which the geo-tag of the Top-1 match was used as the predicted location.

\section{Results and Analysis} \label{sec:results}
\subsection{Overall Quantitative Results}
Table~\ref{tab:qualitative_results} presents the comprehensive performance of all evaluated models across ShipaiVillage dataset. 

{\renewcommand{\arraystretch}{1.6}
\begin{table}[]
\caption{Model performance comparison on ShipaiVillage dataset.}
\label{tab:qualitative_results}
\resizebox{\columnwidth}{!}{%
\begin{tabular}{c|c|cc}
\Xhline{4\arrayrulewidth}
\multirow{2}{*}{\textbf{Methods}} & \multirow{2}{*}{\textbf{Feature Dim.}} & \multicolumn{2}{c}{\textbf{ShipaiVillage Dataset}}                          \\ \cline{3-4} 
                                  &                                        & \multicolumn{1}{c|}{Average Error Distance (m)} & Median Error Distance (m) \\ 
                                  \Xhline{2\arrayrulewidth}
SFRS        & 512  & \multicolumn{1}{c|}{125.025}         & 126.749        \\
CosPlace    & 2048 & \multicolumn{1}{c|}{103.211}         & 80.585         \\
Conv-AP     & 512  & \multicolumn{1}{c|}{104.794}         & 87.855         \\
MixVPR      & 4096 & \multicolumn{1}{c|}{34.877}          & 1.473          \\
EigenPlaces & 2048 & \multicolumn{1}{c|}{85.838}          & 40.282         \\
CricaVPR    & 4096 & \multicolumn{1}{c|}{30.007}          & 1.520          \\
SuperVLAD   & 3072 & \multicolumn{1}{c|}{23.298}          & 1.402          \\
SALAD       & 8448 & \multicolumn{1}{c|}{\textbf{19.510}} & \textbf{1.310} \\ 
\Xhline{4\arrayrulewidth}
\end{tabular}%
}
\end{table}
}

Overall, modern deep aggregation methods significantly outperform earlier approaches in this challenging environment. Traditional methods such as SFRS, CosPlace, and Conv-AP exhibit large localization errors, with AED values exceeding 100 m. These results indicate limited robustness to severe visual clutter, dynamic occlusions, and low illumination conditions commonly observed in urban villages. In contrast, recent state-of-the-art models demonstrate substantial improvements. MixVPR, CricaVPR, and SuperVLAD reduce AED to the range of 23–35 m, while achieving meter-level MED values. Among these methods, SuperVLAD shows stronger robustness, achieving both low AED (23.298 m) and MED (1.402 m), indicating fewer catastrophic failures. SALAD consistently achieves the best performance across all metrics. It attains the lowest AED (19.510 m) and MED (1.310 m), demonstrating superior robustness to extreme appearance similarity, occlusion, and structural repetition.

\begin{figure}[t]
    \centering
    \includegraphics[width=\linewidth]{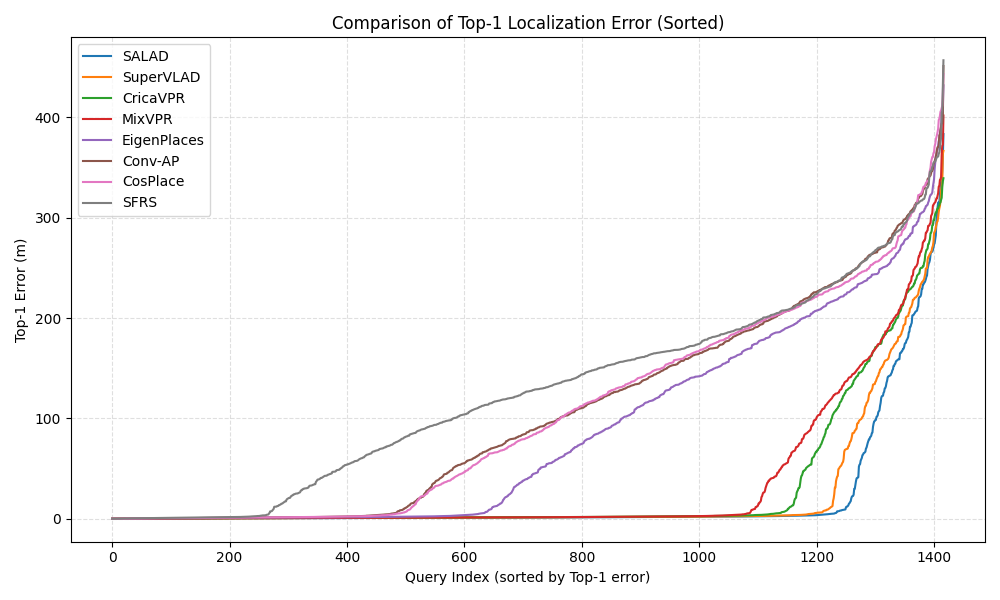}
    \caption{The comparison of the Top-1 localization error distribution of each method.}
    \label{fig:comparison_localization}
\end{figure}

Fig.~\ref{fig:comparison_localization} shows the sorted Top-1 localization error distribution for all evaluated methods. For many queries, most models are able to produce relatively small localization errors, indicating that distinctive visual cues are present in those scenes. As the level of visual ambiguity increases, the differences between models become more apparent. Methods such as SuperVLAD, CricaVPR, and SALAD show more favorable error distributions, with a larger proportion of queries falling within lower error ranges, suggesting stronger robustness across varying levels of scene complexity. In contrast, models like SFRS exhibit a heavier spread toward higher-error regions, reflecting their vulnerability to low-texture or visually repetitive environments. Notably, the extended tail of large errors across all models highlights the inherent difficulty of localizing images in dense urban-village environments.

In summary, the quantitative evaluation demonstrates that state-of-the-art image geo-localization models can achieve meter-level localization accuracy for a substantial portion of queries in urban villages, yet performance degradation under hard cases remains non-negligible. This underscores both the feasibility of visual geo-localization in dense urban informal settlements and the necessity of dedicated benchmarks and models tailored to such environments.

\begin{figure}[t]
    \centering
    \includegraphics[width=\linewidth]{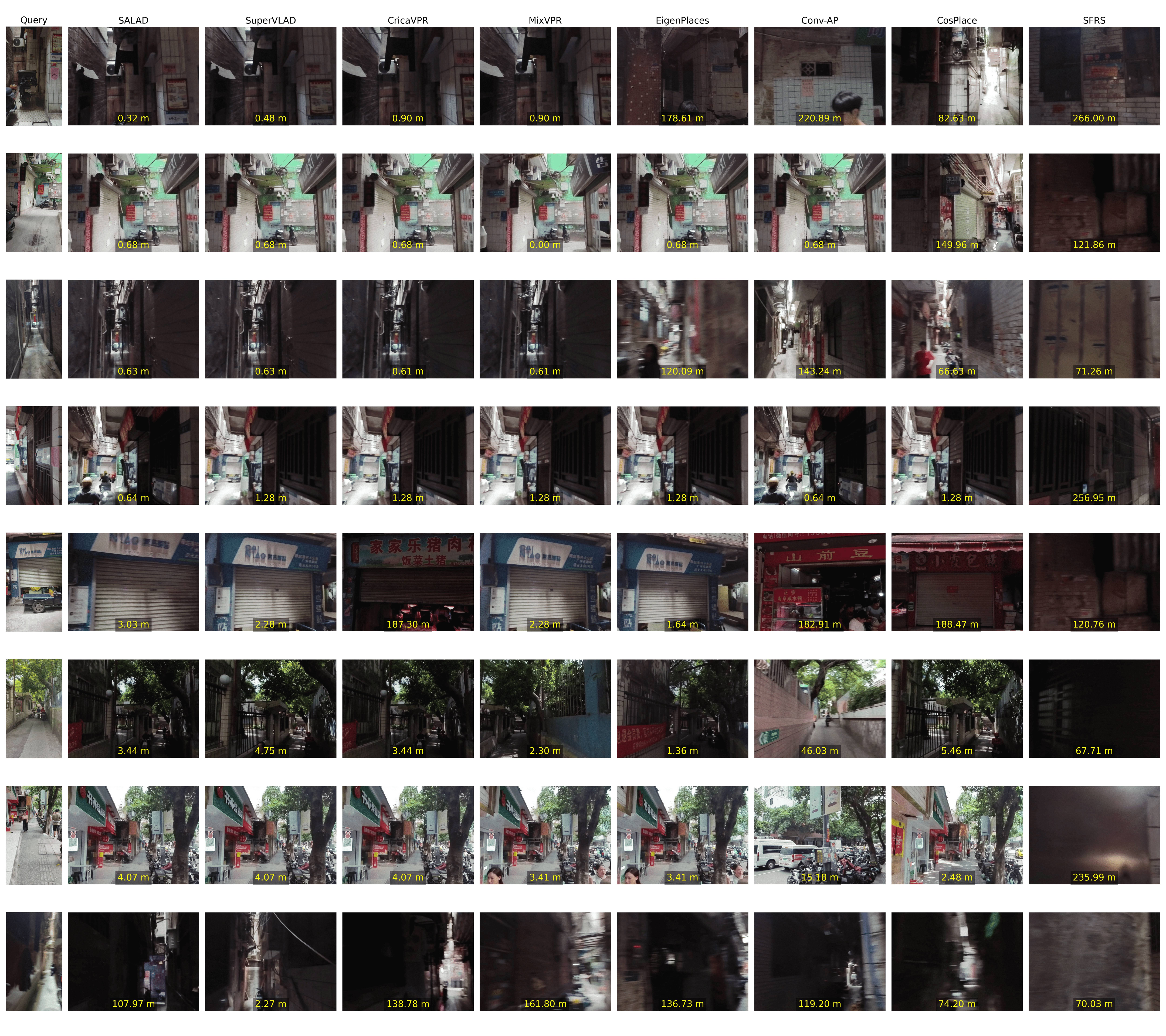}
    \caption{The qualitative results for typical challenging queries in ShipaiVillage dataset.}
    \label{fig:qualitative_result}
\end{figure}

\subsection{Qualitative Results of Challenging Scenarios}
Beyond numerical metrics, we also analyze the qualitative retrieval results on the ShipaiVillage dataset, which contains dense alleyways, repetitive structures, and highly constrained viewpoints.

Fig.~\ref{fig:qualitative_result} presents representative Top-1 retrieval examples along with the corresponding localization errors across a variety of challenging scenarios. For scenes containing distinctive visual cues like recognizable storefront signage, color-contrasting shop shutters, or unique wall textures, most methods are able to retrieve visually and spatially consistent reference images. In these cases, models such as SALAD, SuperVLAD, and CricaVPR typically achieve meter-level localization accuracy, indicating effective exploitation of salient visual cues. 

Conversely, errors increase significantly in narrow corridors where different locations share nearly identical lighting conditions, wall textures, or shuttered shopfronts. In these cases, even high-performing models may retrieve images captured from different but visually similar alleys, resulting in errors spanning tens to hundreds of meters.

Overall, these qualitative findings align with the quantitative results and underscore the challenges posed by ShipaiVillage dataset. The visually repetitive and heavily occlusion nature of urban villages introduces unique difficulties, emphasizing the need for more discriminative visual representations or additional spatial context to achieve reliable geo-localization in dense urban informal settlements.

\begin{figure}[t]
    \centering
    \includegraphics[width=\linewidth]{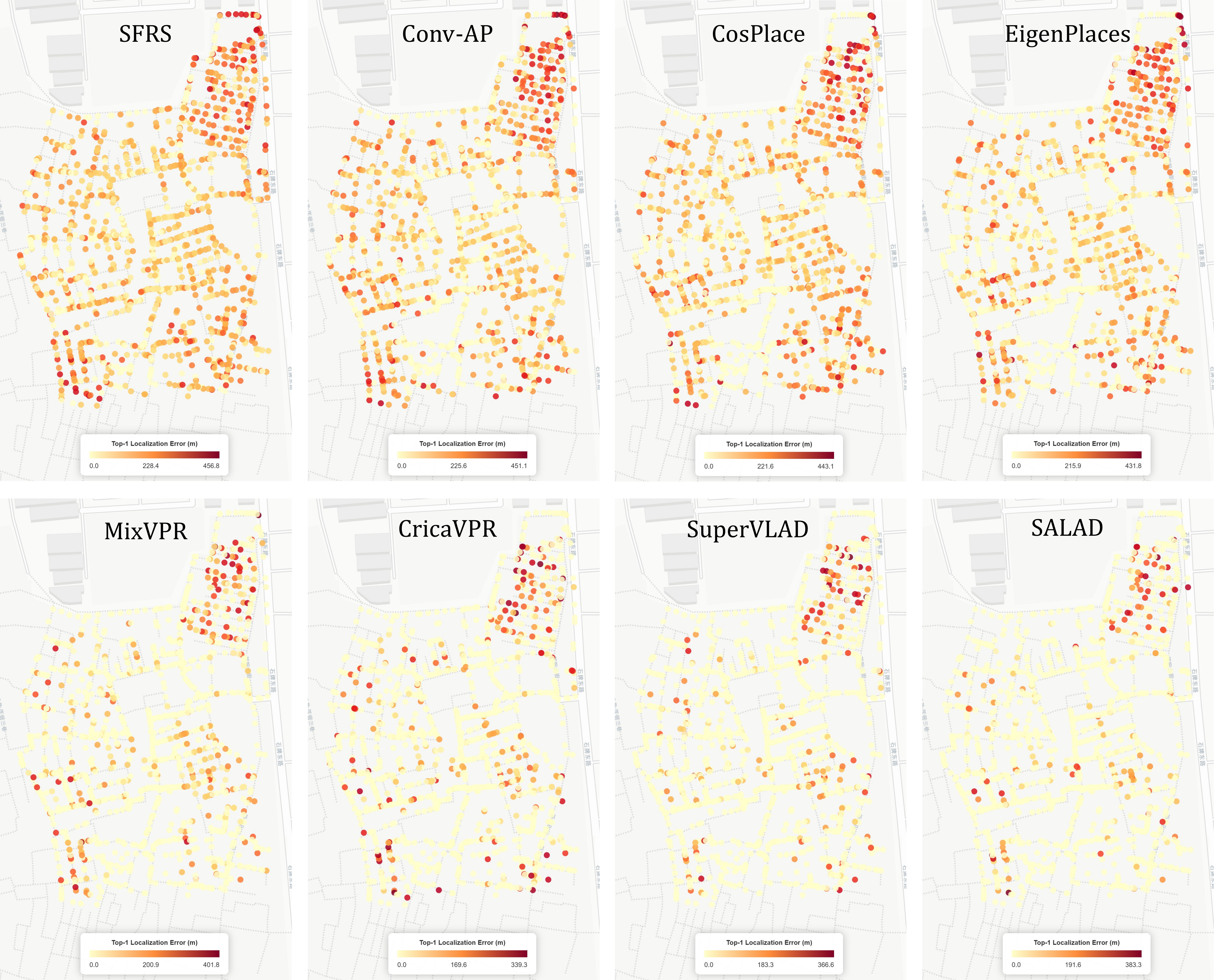}
    \caption{Spatial patterns of Top-1 localization errors for different methods in dense urban-village environment.}
    \label{fig:heatmap}
\end{figure}

\subsection{Spatial Analysis of the Geo-localization Performance}
The spatial patterns of Top-1 localization errors for different methods across the ShipaiVillage dataset are visualized in Fig.~\ref{fig:heatmap}, providing insight beyond aggregate metrics by revealing how errors are distributed within the urban-village environment. Each point corresponds to a query location, with color indicating the magnitude of localization error.

As shown in the figure, SFRS, Conv-AP, CosPlace, and EigenPlaces exhibit widespread high-error regions, particularly along narrow alleys and densely built blocks. These patterns indicate limited robustness to visual repetition and severe occlusion, which are prevalent in urban villages. Errors are spatially dispersed rather than localized, suggesting frequent failures at the pedestrian scale. In contrast, MixVPR, CricaVPR, SuperVLAD, and SALAD substantially reduce the extent of high-error regions. Their error hotspots are more spatially concentrated and mainly appear in particularly challenging areas with homogeneous facades or low-light conditions. This is consistent with their improved localization precisions.

\subsection{Failure Case Analysis}
\begin{figure}[h]
    \centering
    \includegraphics[width=\linewidth]{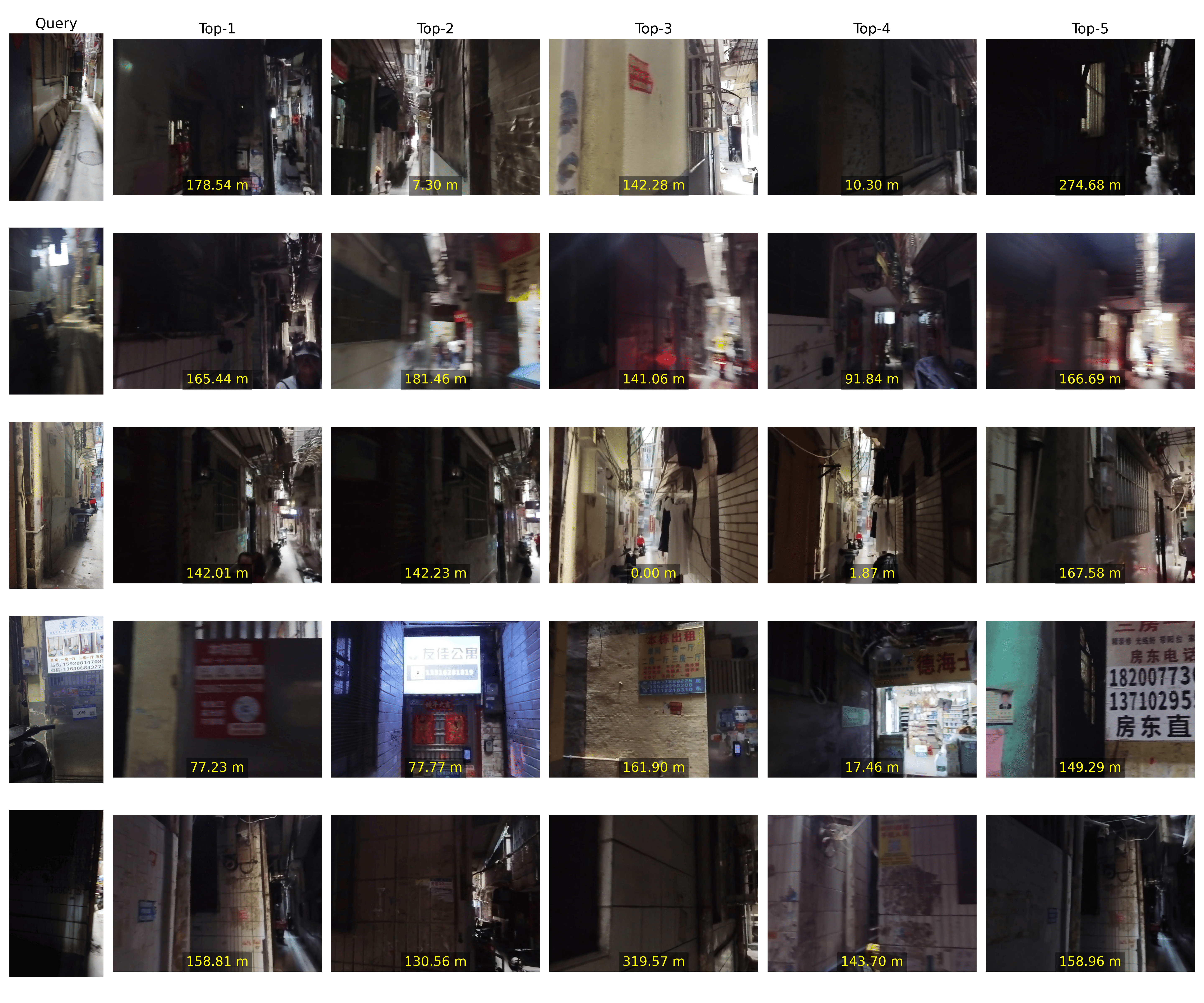}
    \caption{Representative failure cases of SALAD model in dense urban-village environment (Top-1 to Top-5 retrieval results).}
    \label{fig:failed_case}
\end{figure}

To further investigate the underlying causes of these spatially clustered errors, we conduct a closer inspection at the instance level by analyzing representative failure cases. In Fig.~\ref{fig:failed_case}, we present representative failure cases of the SALAD model on the ShipaiVillage dataset, where the Top-1 to Top-5 retrieval results reveal persistent large localization errors for challenging queries. These examples provide deeper insight into the limitations of current state-of-the-art image geo-localization methods in dense urban-village environment.

As illustrated in the figure, a dominant failure mode occurs in narrow alleyways with highly repetitive visual patterns. Many geographically distinct locations share similar wall textures, tiled surfaces, shuttered storefronts, and illumination conditions. In these cases, SALAD retrieves visually similar matches that are in fact far apart spatially. This indicates that although the model effectively captures global appearance similarity, fine-grained local cues that distinguish adjacent alleys are not sufficiently emphasized in the learned embedding.

Another limitation is the insufficient exploitation of textual semantic information. Many failure examples contain informative cues such as shop signs, rental advertisements, or directional notices. However, these textual elements are treated purely as visual patterns rather than semantic entities. As a result, images with different textual meanings but similar visual layouts are often confused, especially in environments where signage styles are repetitive across different streets.

Robustness to image quality degradation also remains a challenge. Several failure cases involve motion blur, strong contrast, or low-light conditions in confined pedestrian environments. Under such conditions, discriminative details are partially lost, and the model relies more heavily on coarse appearance statistics, increasing ambiguity and error magnitude.

In addition, these examples suggest that street-level structural details are not explicitly encoded. Humans can often distinguish visually similar alleys using subtle spatial cues, such as the relative positions of doors, pipes, cables, or bends in the corridor. However, such geometric relationships are difficult to capture, leading to confusion between structurally different but visually similar locations.

In summary, the failure case analysis reveals that while SALAD offers improved robustness compared to prior methods, reliable geo-localization in dense urban villages remains challenging. These observations point to several promising research directions, including integrating text-aware or vision–language representations, enhancing the use of local and structural cues, and improving robustness to low-light and motion-blurred imagery. Incorporating temporal context from image sequences or multimodal information may further help reduce catastrophic failures in highly ambiguous urban-village environments.

\section{Conclusion} \label{sec:conclusion}
This paper investigates image geo-localization in urban villages, a representative yet underexplored class of dense urban environment where conventional positioning systems like GPS often fail. Using a typical urban village Shipai in Guangzhou as a case study, we construct a real-world dataset with a low-cost data collection pipeline and systematically evaluate state-of-the-art retrieval-based localization methods at the pedestrian scale. The proposed pipeline is designed to be easily transferable to other dense and GPS-degraded urban areas, enabling scalable data collection in similarly complex environments. Quantitative and spatial analyses show that recent models achieve substantial improvements and meter-level median accuracy for many queries, while persistent failure cases remain strongly correlated with visually repetitive and low-illumination areas. These findings demonstrate both the practical potential and current limitations of visual-based geo-localization in dense urban-village environments. 
In the future, promising directions worth further exploration include integrating text-aware or vision–language models, leveraging local and structural cues, and improving robustness under low-light or motion blur. Incorporating temporal context from image sequences or multimodal data could further mitigate failures in ambiguous environments.

Overall, this work provides a dedicated benchmark and empirical insights that bridge image geo-localization research with real-world human mobility and emergency-response needs, and we hope it will inspire further development of robust and human-centered localization solutions for complex urban settings and for social good.

\begin{credits}
\subsubsection{\ackname}
This work is supported by the Guangdong Provincial Project (No. 2023QN10H717), the Guangzhou-HKUST(GZ) Joint Funding Program (No. 2025A\\03J3639), and AI Research and Learning Base of Urban Culture Project (No. 2023WZJD\\008).
\end{credits}

\bibliographystyle{splncs04}
\bibliography{references}
\end{document}